\documentclass[smallextended,natbib,runningheads]{svjour3}
\journalname{Annals of the Institute of Statistical Mathematics}
\smartqed  
\usepackage{graphicx,amsmath,amssymb,natbib,mathtools,tabularx,xcolor,hyperref}
\usepackage[cal=boondoxo]{mathalfa}
\mathtoolsset{showonlyrefs,showmanualtags}
\usepackage{xr}
\newcommand{\AIC}{\ensuremath{\textrm{AIC}}}
\newcommand{\QIC}{\ensuremath{\textrm{QIC}}}
\newcommand{\BIC}{\ensuremath{\textrm{BIC}}}

\newcommand{\MLE}{\hat{  {\theta}}}
\newcommand{\KLstat}{{d}}
\newcommand{\KLD}{D}
\newcommand{\Comp}{{ \cal K}}
\newcommand{\dimK}{K}

\newcommand{\expectation}[2]{\underset{#1 \vert #2}{\mathbb{E}} \,}
\newcommand{\expect}[1]{\underset{#1}{\mathbb{E} }\,}

\newcommand{\idea}[1]{\subsection{#1}}
\newcommand*{\Scale}[2][4]{\scalebox{#1}{$#2$}}%
\newcommand*{\inliner}[1]{{\smash{\Scale[0.9]{#1}}}}
\newcommand{\bm}{}

\begin{document}

\title{Information-based inference for singular models and finite sample sizes
\thanks{This work was supported by NSF-MCB-1243492}}
\subtitle{A frequentist information criterion}


\author{Colin H. LaMont        \and
       Paul A. Wiggins 
}


\institute{C. H. LaMont \at
           \email{lamontc@uw.edu}         
         \and
           P. A. Wiggins \at
           \email{pwiggins@uw.edu}\\
           Departments of Physics, Bioengineering and Microbiology\\
           University of Washington, Box 351560.\\ 
           3910 15th Avenue Northeast, Seattle, WA 98195, USA
}

\date{Received: date / Revised: date}

\maketitle
\pagebreak
\begin{abstract}
In the information-based paradigm of inference, model selection is performed by selecting the candidate model with the best estimated predictive performance. 
The success of this approach depends on the accuracy of the estimate of the predictive complexity.
In the large-sample-size limit of a regular model, the predictive performance is well estimated by the Akaike Information Criterion (\AIC). However, this approximation can either significantly under or over-estimating the complexity in a wide range of important applications where models are either non-regular or finite-sample-size corrections are significant.
We introduce an improved approximation for the complexity that is used to define a new information criterion:
the  {\it Frequentist Information Criterion} (QIC).  QIC extends the applicability of information-based inference to the finite-sample-size regime of regular models and to singular models. We demonstrate the power and the comparative advantage of QIC in 
a number of example analyses.
\keywords{Model selection \and Information criteria \and AIC \and Singular models \and Finite sample size \and Hypothesis testing \and Frequentist}
\end{abstract}

\section{Introduction}
Model selection is a central problem in statistics. In the information-based paradigm of inference, models are selected to maximize the expected predictive performance.  The canonical implementation of information-based inference is the minimization of the Akaike Information Criterion (AIC), an estimate for the (minus) predictive performance  \citep{akaike1773,BurnhamBook}. 
Although  it  has enjoyed significant success, AIC is biased in many important applications.
Model singularity, i.e.~the absence of a one-to-one correspondence between model parameters and distribution functions, can make the bias extremely large and result in the catastrophic failure of model selection, as described below. 
There are three important and related mechanisms of failure: 
	(i) finite-sample-size corrections, 
	(ii) model singularity and
	(iii) model-training-algorithm dependence.  
In the course of our own analyses of biophysical and cell biology data, we frequently encounter all three phenomena. The goal of this paper is to propose a refinement to the information-based approach that overcomes these limitations.

We begin by studying the predictive complexity that plays a critical role in the mechanism of failure of AIC. We compute the exact predictive complexity of models to study its phenomenology  and  dependence on the parameters of the generative model.
We discover that the AIC approximation for the complexity can  significantly under or over-estimate the complexity, leading to pathological over-fitting or under-fitting in model selection problems. We find that parameter unidentifiability (i.e.~model singularity), sample size, fitting algorithm and parameter manifold geometry can all play a critical role in determining the model complexity.


In real analyses, the true distribution is unknown and therefore the complexity must be approximated.
Our exploration of the true complexity motivates a new approximation for the complexity: the {\em frequentist complexity}. In this approximation, we assume the model of interest is the generative model at the estimated parameters.  The frequentist complexity is not a universal function of model dimension and sample size.
Instead it naturally adapts to the likelihood function, model training algorithm and sample size.
We propose an improved information criterion based on this new frequentist complexity: the {\it Frequentist Information Criterion} (QIC).

For regular models in the large-sample-size limit, QIC is equal to AIC.
Away from this limit, there can be large mismatches between the QIC and AIC. For instance, for models with large multiplicity, QIC can be much larger than AIC. For \textit{sloppy} models \citep{Machta:2013hl}, QIC can be much smaller than AIC.
It is essential to note that QIC is still biased (since the true distribution is not know) but this bias is nearly always much smaller than the AIC estimate of the complexity and, as a result, QIC  outperforms AIC (and other information criteria). QIC also outperforms other predictive methods in many contexts. 
To demonstrate this improved performance, we present three example analyses in Section \ref{sec:app} that highlight specific advantages QIC over other methods.


\section{Information-based inference}
The goal of statistical modeling in this discussion is to approximate the unknown true distribution function $p$ which generated an observed  dataset:
\begin{equation}
\smash{x \equiv (x_1 \hdots x_N)},
\end{equation}
of sample size $N$.  We will use $X$ (instead of $x$) when we interpret $X$ as random variables.
The model ${\cal m}$ consists of a parameterized candidate probability distribution $\inliner{q( x | {\theta}^{\cal m} )}$, called the likelihood,  with parameters ${\theta}^{\cal m}$ and an algorithm for training the model $\hat{\theta}^{\cal m}$ \citep{mccullagh2002}.  The dependence of all quantities on the model $\cal m$ will be implicit, except where we make explicit comparisons between competing models. 
We will work predominantly in terms of Shannon information, defined:
\begin{equation}
h(x|{ \theta}) \equiv - \log q(x|{\theta}),
\end{equation}
where $h$ is the base-e message length (in nats) required to encode $x$ using distribution $q(\cdot|{\theta})$.
The output of a model-training algorithm, trained on measurements $x$, is a set of parameters $\inliner{\hat{  \theta}_x = \hat{  \theta}(x)}$. 
The methods that we explore apply to any model-training algorithm.  For concreteness, we focus on models trained using maximum-likelihood parameter estimation. The Maximum-Likelihood-Estimate (MLE) of the parameters $\hat{  \theta}_x$ is found by maximizing (minimizing) the likelihood $q(x|{\theta})$ (information $h$) with respect to the parameters ${  \theta} \in { \Theta}$.

It will be convenient to view both the true model and the candidate model parameter space ${\bm \Theta}$ as embedded in a higher dimensional space ${\bm \Phi}$, so ${\bm \Theta} \subseteq {\bm \Phi}$ with the true model parameterized by $\phi \in {\bm \Phi}$. We define the expected excess information loss, \textit{i.e.} the KL-divergence: 
\begin{align}
\KLD(\phi || \theta) \equiv \expectation{X}{\phi} h(X|\theta) - h(X|\phi).
\end{align}
The information loss, the empirical estimator for the KL-divergence is given by:
\begin{align}
\KLstat_x(\phi || \theta) \equiv h(x|\theta) - h(x|\phi).
\end{align}
$\KLD$ and $\KLstat$ act as directed distance functions and define a geometry for the parameter  space of the model termed the {\em statistical manifold} \citep{amari2016information,barndorff1997yokes,komaki1996asymptotic}.
For small perturbations around the true parameters, the KL-divergence can be computed using the Fisher information:
\begin{align}
I(\theta) = \lim_{\theta' \rightarrow \theta} \partial_{\theta'} \otimes \partial_{\theta'} \KLD(\theta || \theta'),
\end{align}
which can be reinterpreted as  the Fisher-Rao metric and defines  a local notion of distance on the manifold \citep{balasubramanian1997,burbea1982entropy}.
 
\subsection{Information criteria} The true distribution is approximated in two steps: (i)  the parameters $\hat{\theta}$ are selected in each model ${\cal m}$ as described above
and (ii) a  model $\hat{\cal m}$ is then selected among a small number of competing models. 
In information-based inference, models are selected to maximize the estimated \textit{predictive performance}. 
Predictive performance of the fitted model parameterized by $\hat{  \theta}_x$ is measured by the cross entropy:
\begin{equation}
H(\phi||\hat{  \theta}_x) \equiv \expectation{X}{\phi} h(X|\hat{  \theta}_x),
\end{equation}
where $X$ has identical structure to the observed data $x$.  The model with the smallest cross entropy is the most predictive model, but $H$ is unknown since $\phi$ is unknown.  

In information-based inference, $H$ is approximated by an \textit{information criterion}. We will use the information $h$ as an empirical estimator. 
Although $\inliner{h(x|{  \theta})}$ is  an unbiased estimator of $\inliner{H(\phi||\theta)}$, $\inliner{h(x|\hat{  \theta}_x)}$ is biased from below:
\begin{equation}
\expectation{X}{\phi} h(X|\hat{  \theta}_X) \le \expectation{X}{\phi} H(\phi||\hat{  \theta}_X).
\end{equation}
$\inliner{h(x|\hat{  \theta}_x)}$ describes in-sample performance, but in-sample and out-of-sample performance are distinct due to the phenomenon of overfitting. 
In the context of nested models\footnote{An important class of models is referred to as {\it nested} \citep{mccullagh2002}.
Lower-dimension model ${\cal m}$ is nested in higher-dimensional model ${\cal n}$ if all candidate distributions in $\inliner{\Theta^{\cal m}}$ are realizable in $\inliner{\Theta^{\cal n}}$.}, this bias in  $\inliner{h(x|\hat{  \theta}_x)}$ cannot be ignored since  $\inliner{h(x|\hat{  \theta}^{\cal m}_x)}$ typically monotonically decreases with model dimension even as the cross entropy $\inliner{H(\phi||\hat{  \theta}^{\cal m}_x)}$ increases. Minimizing  $\inliner{h(x|\hat{  \theta}^{\cal m}_x)}$ with respect to ${\cal m}$ would lead to the selection of the most complex model.

To select the  model with optimum predictive performance, we must correct the bias of the cross-entropy estimator $h(x|\hat{  \theta}_x)$. 
This bias is defined:
\begin{eqnarray}
{\cal K} \equiv  \expectation{X}{\phi} \left\{ H(\phi || \hat{\theta}_X) - h(X| \hat{\theta}_X) \right\}, \label{eq:comp}
\label{eqn:comp}
\end{eqnarray}
but for the purposes of computation, it is often convenient and more computationally efficient to re-write the bias in terms of the KL divergence:
\begin{eqnarray}
{\cal K} =  \expectation{X}{\phi} \left\{ \KLD(\phi || \hat{\theta}_X) - \KLstat_X(\phi || \hat{\theta}_X) \right\}.
\label{eqn:comp}
\end{eqnarray}
${\cal K}$ is called the  predictive complexity, or \textit{complexity} in the interest of brevity. 
The complexity can be understood  intuitively as the \textit{flexibility} of the model in fitting data $x$.

By construction, an unbiased estimator of the cross entropy $\inliner{H(\hat{  \theta}_x)}$ is:
\begin{equation}
{\rm IC}(x) = h(x|\hat{  \theta}_x)+{\cal K},
\label{eqn:IC}
\end{equation}
which is called an \textit{information criterion}. 
The first term, the minimum information $h$, measures the \textit{goodness-of-fit} of the model and typically decreases with model complexity (or dimension). The second term, the \textit{complexity}, is a penalty that represents expected information loss due to the statistical variation of the parameter values fit to the training set $x$. As the model dimension increases, so does the complexity, while 
the information $h$ decreases with model dimension.
As a consequence of these competing imperatives (improving the fit while minimizing the model complexity), 
the information criterion has a minimum with respect to model dimension corresponding to the estimated optimally predictive model. 

\subsection{The Akaike Information Criterion} 

Although neither the complexity (Eqn.~\eqref{eqn:comp}) nor the information criterion (Eqn.~\eqref{eqn:IC})  can be computed if the true parameters $\phi$ are unknown, in practice the $\phi$ dependence vanishes asymptotically. In the large-sample-size limit of a regular model, a surprisingly simple expression is derived for the complexity:
\begin{equation}
{\cal K} = K+{\cal O}(N^{-1}), \label{eqn:AICcomp}
\end{equation}
where the model dimension  is $K \equiv \dim \Theta$. 
This complexity approximation can be understood as the leading-order contribution to a perturbative expansion of the complexity in inverse powers of the sample size $N$. 
Using Eqn.~\eqref{eqn:AICcomp}, we can write the well-know Akaike Information Criterion (AIC)\footnote{Historically, AIC was defined as twice Eqn.~\eqref{eqn:AIC} for consistency with the deviance \citep{BurnhamBook}. There is no significance to this multiplicative factor. }:
\begin{equation}
{\rm AIC}(x) = h(x|\hat{  \theta}_x)+K, \label{eqn:AIC}
\end{equation}
which does not depend on  (i) the true distribution $\phi$, (ii) the detailed functional form of the candidate models $\inliner{q(x|{  \theta})}$, (iii) the data structure or (iv) the sample size $N$.

Although the AIC information criterion has been successfully applied in many problems,  the AIC approximation for the complexity can fail in many unexceptional contexts. For instance,  the $\inliner{{\cal O}(N^{-1})}$ correction may not be small at finite sample size. 
Alternatively, the structure of the model can cause AIC to fail. For instance, a parameter ${  \theta}$ is called \textit{unidentifiable} if $\inliner{q(\cdot|{  \theta})=q(\cdot|{  \theta}')}$ for $\inliner{{  \theta} \ne {  \theta}'}$.
If a model includes unidentifiable parameters, the model is called \textit{singular}, as opposed to a \textit{regular} statistical model \citep{watanabe2009}.  For singular models, AIC fails at all samples sizes. As our examples in Sec.~\ref{sec:app} will illustrate, both these mechanisms of failure naturally arise in many analyses.

\section{Complexity Landscapes}

\label{sec:Land}
To study the phenomenology and investigate novel approximations for the complexity, we compute it for realizable models as a function of the true parameter $\theta$. We will find that although the AIC complexity is correct in the large-sample-size limit of a regular model, there can be significant deviation from this approximation at finite sample size, in singular models, and as a result of parameter-space constraints.

\subsection{The finite-sample-size complexity of regular models}
\label{sec:constantComp}
In general, the complexity will depend on both the sample size $N$ and the true parameter $\theta$. However, statistical models with symmetries can lead to a complexity independent of the true parameter. For instance, consider a family of distributions:
\begin{equation}
q(x|\theta) = \textstyle C_\alpha\, \lambda^{1/\alpha}\, e^{-\lambda| x|^\alpha}, \label{eqn:gausslaplace}
\end{equation}
with parameters $\theta = (\lambda, \alpha)$ and support $\lambda,\alpha \in {\mathbb R}_+$ and normalization:
\begin{equation}
C^{-1}_\alpha \equiv \Gamma(1+\alpha^{-1})\times \begin{cases} 2, & x \in {\mathbb R} \\
                  1, & x \in {\mathbb R}_+.
                  \end{cases}
\end{equation}  
This family includes the exponential ($\alpha = 1$, $x \in {\mathbb R}_+$), the centered-Gaussian ($\alpha = 2$, $x \in {\mathbb R}$), the Laplace ($\alpha = 1$, $x \in {\mathbb R}$) and the uniform ($\alpha \rightarrow \infty$) distributions.  The transformation of this distribution under dilations on $x$ implies the complexity must be independent of $\lambda$. 
In the Appendix Sec.~\ref{app:bregmann}, we derive a general result for exponential-family models. This problem is a special case of that expression. The complexity for unknown $\lambda$ and known $\alpha$ is:
\begin{equation}
{\cal K} = \textstyle \frac{N}{N-\alpha}\ \ \ \ {\rm for} \ \ \ \ \ N>\alpha, \label{eqn:alphacomp}
\end{equation}  
as shown in Appendix Sec.~\ref{sec:modgauss}.

\begin{figure*}
  \centering
    \includegraphics[width=.5\textwidth]{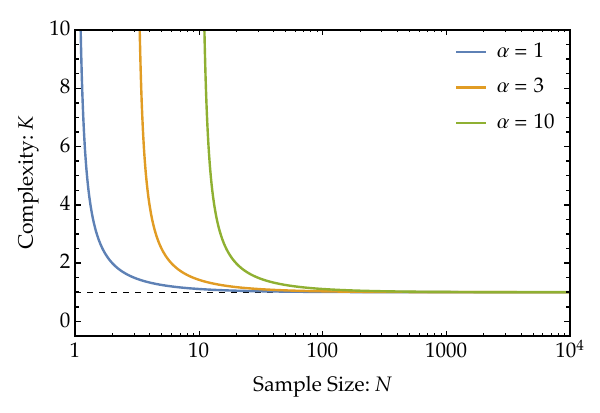}
      \caption{
       {\bf Complexity at finite sample size.} Although AIC estimate accurately estimates the large-sample-size limit of the complexity of regular models, there can be significant finite-sample-size corrections. For instance, the modified-center-Gaussian model has a significantly larger complexity than the AIC limit for small $N$. In fact the complexity diverges for $N\le \alpha$, implying that the model has insufficient data to make predictions.
       \label{complexitylandscapeFinite}}
  \end{figure*}

The complexity of this centered-modified Gaussian family is equal to one asymptotically ($N\rightarrow \infty$) but can significantly diverge from this AIC limit at finite sample size $N$, as shown in Fig.~\ref{complexitylandscapeFinite}. The finite-sample-size correction is particularly large for large values of the exponent $\alpha$. In this regime, the MLE algorithm tends to strongly overestimate the fit to the data. In fact, the complexity is infinite in the uniform distribution limit ($\alpha \rightarrow \infty$) where a Bayesian approach, which hedges over parameter $\lambda$, is required to give acceptable predictive performance at any sample size $N$. (See Ref.~\cite{Correspondence2018} for more information.)
%
%

\begin{figure*}
  \centering
    \includegraphics[width=1.0\textwidth]{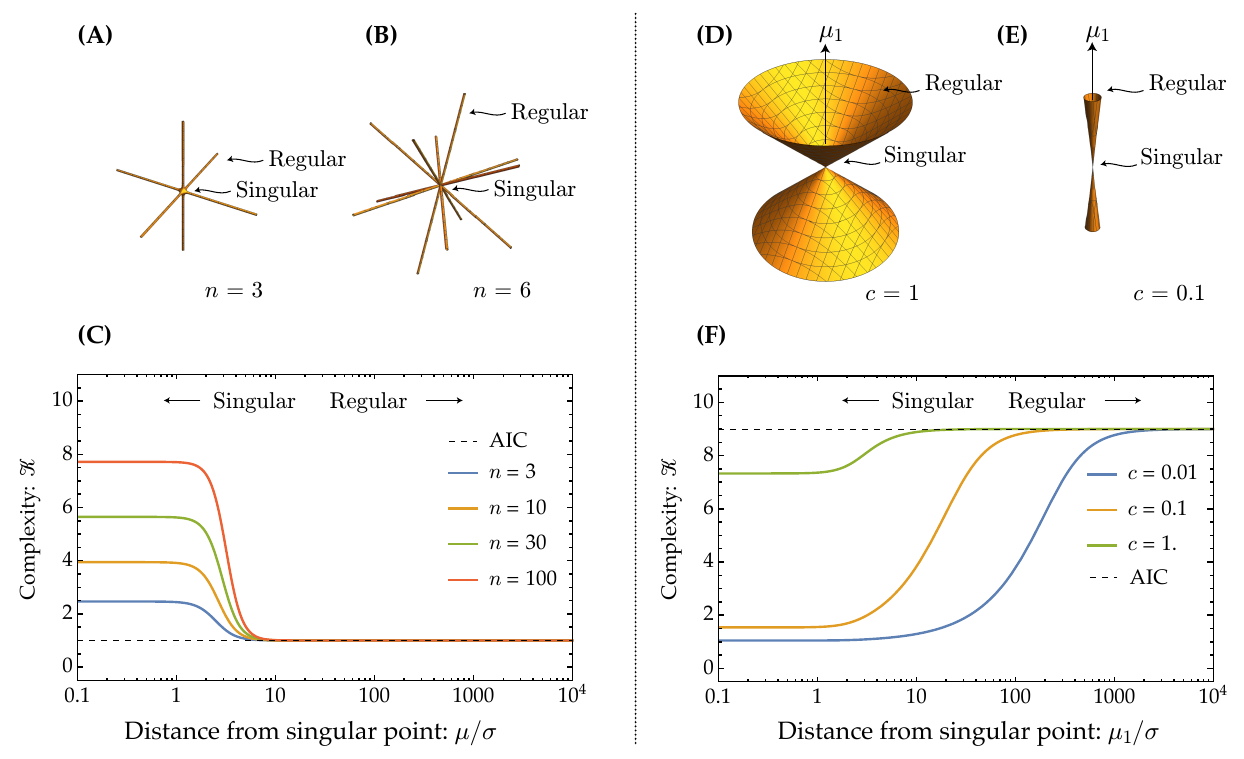}
      \caption{
       {\bf Complexity landscapes in singular models. AIC underestimates the complexity in the component selection model. Panel A-B:} Schematic sketches of the geometry of parameter space for two different multiplicity values: $n=3$ and $n = 6$. \textbf{Panel C:} The AIC estimate ${\cal K} = 1$ (dashed line) matches the true complexity far from the singular point ($|\mu/\sigma| \gg 0$). Close to the singularity ($|\mu/\sigma| \approx 0$), the true complexity is much larger than the AIC estimate. The complexity grows with the number of means $n$ due to \textit{multiplicity}. 
      {\bf AIC overestimates the complexity in the n-cone model. Panel D-E:} Schematic sketches of parameter space for 
      a wide cone ($c=1$) and a needle-like cone ($n=0.1$). \textbf{Panel F:} 
      For $n=10$ dimensions, the AIC estimate ${\cal K} = n-1$ (dashed line) matches the true complexity far from the singular point ($|\mu_1/\sigma| \gg 0$). Close to the singularity ($|\mu_1/\sigma| \approx 0$), the true complexity is much smaller than the AIC estimate. The complexity shrinks for small cone anles ($c\rightarrow 0$) since the cone geometry is needle-like with effectively a single degree for freedom ($\mu_1$).
\label{complexitylandscape1}}
  \end{figure*}%

\begin{figure*}
  \centering
    \includegraphics[width=1.0\textwidth]{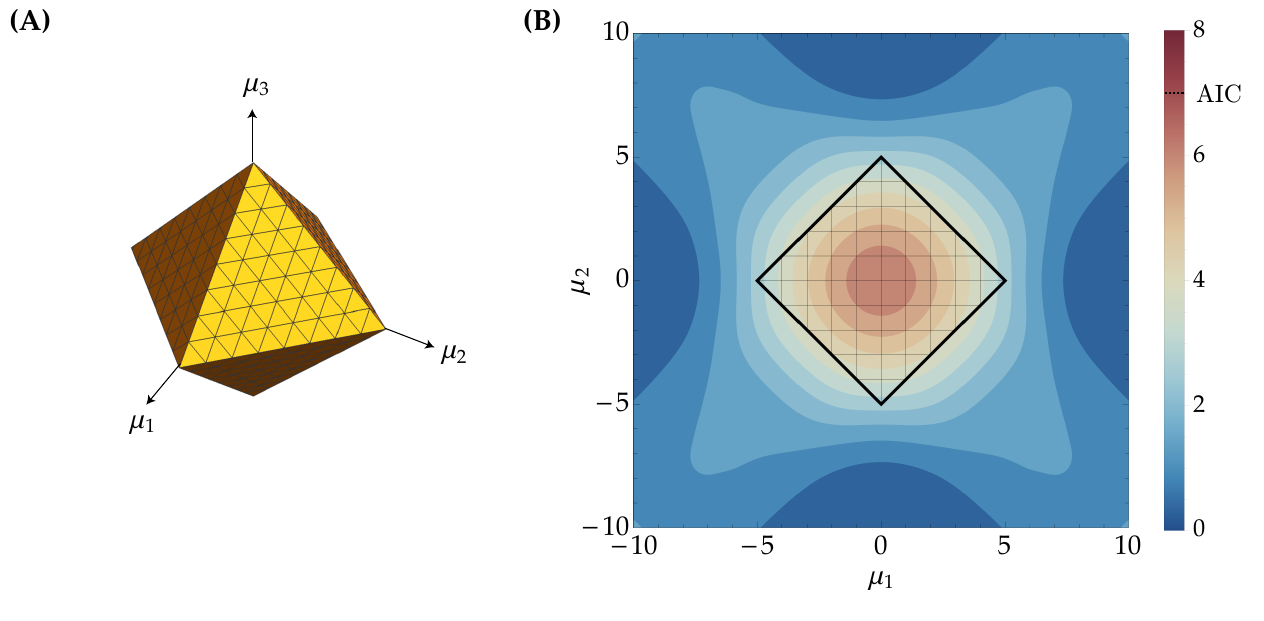}
      \caption{
 \textbf{Complexity of $L_1$-constrained model.  Panel A:} Schematic sketch of a slice of the seven-dimensional parameter space. Parameter values satisfying the $L_1$ constrain lie inside the simplex.
	  {\bf Panel B:} Complexity as a function of the true parameter value $\vec{\mu}= (\mu_1,...,\mu_7)$. (Only a slice representing the $x$-$y$ plane is shown.) The black-hatched region represents parameter values satisfying the constraints. The $L_1$ constraint significantly reduces the complexity below the AIC estimate $\cal K = 7$. The complexity is lowest outside the boundaries of the simplex where the constraints trap MLE parameter estimates and reduce statistical fluctuations.
    \label{complexitylandscape3}}
  \end{figure*}%

\subsection{Singular models}

Singular models have parameter unidentifiability that cannot be removed by coordinate transformation. These models can show very large deviations from the AIC complexity at all sample sizes in the vicinity of the singular point in parameter space. The deviation can either significantly increase or decrease the complexity as we will illustrate with two closely related examples. This singular class of models are common place in many analyses, especially in the context of nested models, and therefore they pose a significant limitation to the more general use of AIC.

To explore the properties of a singular model, consider the following simple example: the component selection model. An $n$-dimensional vector of observations $\vec{x}\in {\mathbb R}^n$ is normally distributed about an $n$-dimensional vector of means ${\vec{\mu}}\in {\mathbb R}^n$ with variance one. The likelihood is:
\begin{align}
q(\vec{x}|{\bm \theta}) &= (2\pi)^{-n/2} \exp[ -\textstyle\frac{1}{2}(\vec{x}-\vec{\mu})^2]. \label{eqn:MDG}
\end{align}
We consider the model where all but one of the components of the vector mean are zero: 
\begin{align}
\vec{\mu} &= (0,...,\mu_i,...,0),
\end{align}
but the identity, $i$, of the non-zero component is unknown as well as the mean $\mu_i \equiv \mu$. The parameters are defined $\theta \equiv (i,\mu)$ with support $\mu \in \mathbb R$ and where the index $i$ is an integer on the interval  $[1,n]$. This model is singular when $\mu = 0$ since the likelihood is independent of $i$. The complexity must be computed numerically (Appendix Sec.~\ref{sec:compSel}) and depends on $\mu$ but is independent of $i$ (permutation symmetry) and is plotted in Fig.~\ref{complexitylandscape1}A. As shown in the figure, there is a large deviation from the AIC complexity in the singular region $\mu = 0$ and the complexity is large compared with the model dimension, irrespective of sample size $N$. Far from the singular point, the complexity is ${\cal K}=1$ which matches the AIC complexity for a single continuous parameter ($\mu$) and the discrete parameter $i$ does not contribute to the complexity in this limit.


To demonstrate that singular models can have reduced complexity relative to AIC, consider the same likelihood function (Eqn.~\eqref{eqn:MDG}), but a different parameter manifold. We constrain the mean $\vec{\mu}$ to lie on the surface of a $n$-cone, defined by the equation:
\begin{equation}
\mu_1^2 c^2 = \sum_{i=2}^{n}\mu_i^2,
\end{equation}
where $\alpha = \tan^{-1} c$ is the angle of the cone. This cone geometry has been previously suggested to represent the fundamental geometry mixture models \citep{hagiwara2001upper,amari2002geometrical}. The model is singular at the vertex of the cone $\mu_1 = 0$. The complexity can be computed analytically (Appendix Sec.~\ref{sec:conecalc}) and is shown in Fig.~\ref{complexitylandscape1}B. Like the previous singular model, there is a large deviation from the AIC complexity at the singular point $\mu_1 = 0$ where the complexity is small ($\cal K \approx 1$) compared with the model dimension ($n-1$), irrespective of sample size $N$. Far from the singular point, the complexity is ${\cal K}\approx n-1$, which matches the AIC complexity for $n-1$ dimensional parameter manifold. 
In general, we expect a strong failure of the AIC approximation in the vicinity of the singularity, but far from the singular point, the AIC approximation applies in the large-sample-size limit.

\subsection{Constrained models}

\label{sec:l1}
A canonical approach to regularizing high-dimensional models are convex constraints, including $L_1$ constrained optimization.  Consider the same likelihood function (Eqn.~\eqref{eqn:MDG}), but with convex constraint:
\begin{align}
\sum_{i=1}^n |\mu_i| &\leq \lambda, \label{eqn:contr}
\end{align}
where $\lambda$ is a constraint chosen by the analyst.
The complexity landscape can be computed numerically (Appendix Sec.~\ref{app:L_1}) and is shown in Fig.~\ref{complexitylandscape3}. As expected, the constraint works to significantly reduce the complexity far below the AIC estimate at finite sample size, especially when the true parameter lies somewhere close or outside the subspace of parameter space that satisfies the constraint (Eqn.~\eqref{eqn:contr}).

\section{Frequentist Information Criterion (QIC)}

In each example discussed in the previous section, we demonstrated a significant mismatch between the AIC complexity and the true complexity. In practice, these corrections are often important since (i) singular model are widespread and (ii)  all real analyses occur at finite sample size. A significant bias in the complexity can lead to failures in model selection and, in the context of recursively-nested singular models, it can lead a catastrophic breakdown in model selection where the selected model dimension grows with sample size indefinitely, irrespective of the generative distribution (\textit{e.g.}~Sec.~\ref{Sec:Neutrino}). Our goal is therefore to develop an improved approximation for the complexity.

Clearly the ideal situation would be to  use the true complexity ${\cal K}({  \phi})$, but ${\cal K}({  \phi})$  depends upon the unknown generative distribution, \textit{i.e.} the unknown parameter $\phi$. To circumvent this difficulty, we propose using a natural approximation in the current context: We approximate $\theta$ with the point estimate $\hat{\theta}_x$ and define the frequentist approximation of the complexity:
\begin{equation}
{\cal K}_{\rm QIC}(x) \equiv {\cal K}(\hat{  \theta}_x). \label{eqn:QICcomp}
\end{equation}
where ${\cal K}({  \theta})$ is the true complexity for data generated by a realizable distribution with parameter ${  \theta}$. We call this a \textit{frequentist approximation} since the \textit{complexity is computed with respect to hypothesized data distributions} in close analogy to the computation of the distribution of a frequentist test statistic.  
Unlike a frequentist test, no \textit{ad hoc} confidence level must be supplied by the analyst. We generically expect the frequenist complexity to depend on (i) the data $x$, (ii) the functional form of candidate models $q$, (iii) the training algorithm and (iv) the sample size $N$. In general, the complexity must be computed numerically, although analytic results or approximations can be used in many models. 
 
 Now that we have defined a novel approximation for the complexity, we can define the corresponding information criterion:
\begin{equation}
{\rm QIC}(x) = h(x|\hat{  \theta}_x)+{\cal K}_{\rm QIC},\label{eqn:QIC}
\end{equation}
which we call the Frequentist Information Criterion (QIC). In analogy to the AIC analysis, the model that minimizes QIC is estimated to have the best predictive performance.

\subsection{Measuring model selection performance}

The QIC approach is to compute an approximate complexity (Eqn.~\eqref{eqn:QICcomp}) in order to construct the information criterion (Eqn.~\eqref{eqn:IC}), an estimator of the cross entropy $\inliner{H(\phi||\hat{  \theta}_x)}$. In simulations, $\phi$ is known. Therefore the estimated complexity can be compared with the true complexity and the information criterion can be compared with the cross entropy $\inliner{H(\phi||\hat{  \theta}_x)}$. 

A more direct metric for the performance of information criteria is the information loss of the selected model $\hat{\cal m}$. The selected model $\inliner{\hat{\cal m}}$ is that which minimizes the information criterion
\begin{equation}
\hat{\cal m}(x) = \arg \min_{\cal m} \mathrm{QIC}^{\cal m} (x) \label{eqn:selected}.
\end{equation}
The expected performance of a selection criterion is the KL divergence averaged over training sets $X$,
\begin{equation} 
\overline{D} \equiv  \expectation{X}{\phi} D(\phi||\hat{\theta}^{\hat{\cal m}(X)}_X), \label{eqn:ModelKL} 
\end{equation}
where $q_X(\cdot)$ is the estimate of $p$, which is the result of the model selection procedure \eqref{eqn:selected}.
The better the performance of the model selection criterion, the smaller the information loss $\overline{D}$. 

\section{Applications of QIC}

\label{sec:app}

In Sec.~\ref{sec:Land}, we described two important contexts in which AIC fails: (i) finite sample size and (ii) in singular models. 
Before considering a formal analysis of the performance of QIC, we explore this criterion in the context of a number of sample problems. 
First, we analyze a problem of modeling the motion of large complexes in the cell in Sec.~\ref{Sec:Subdiffusion}. In this problem, finite sample size plays a central role in the choice of models when we compare two models with the same dimension. As expected, QIC outperforms AIC. In the next analysis in Sec.~\ref{Sec:Neutrino}, we analyze a Fourier Regression problem. In this analysis, we fit the data using two different algorithms, one of which is singular. In the analysis of the singular model, there is a catastrophic failure of AIC where the dimension of the AIC-selected model is much larger than the optimally predictive model due to the large size of the true complexity relative to the AIC estimate. Again, we demonstrate that QIC gives a good approximation for the complexity, both in the context of the singular and regular models. 
In the final example in Sec.~\ref{Sec:expDecay}, we analyze a singular model in which the complexity is significantly smaller than the model dimension. As expect, QIC outperforms AIC in this context as well.

\subsection{Small sample size and the step-size analysis}
\label{Sec:Subdiffusion}

\begin{figure*}
  \centering
    \includegraphics[width=.60\textwidth]{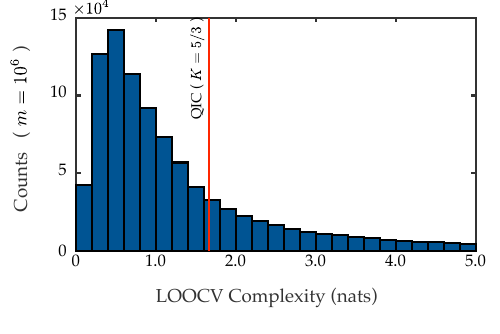}
      \caption{
     	 {\bf Monte carlo histogram of the LOOCV procedure:} A histogram of $10^6$ simulations of the effective cross-validation complexity $\Comp_\mathrm{CV} (X)$ for the normal model with unknown variance $\alpha = 2$ compared to the QIC result ${\cal K}= \frac{5}{5-2}$ for ($N = 5$). The lower variance of the QIC complexity often results in better model selection properties, especially at low sample sizes relative to cross-validation. \label{LOOCV} }
\end{figure*}

In this section, we explore the small-sample-size limit in the context of a problem with two competing models of the same dimension but different true complexities. Inspired by our recent experimental work \citep{lampo2017cytoplasmic,stylianidoustrong}, we  model the step-size distribution of large protein complexes in the cytoplasm undergoing stochastic motion. In this problem, individual complexes can only be tracked over a short interval of $N_t$ steps. Although many trajectories can be captured ($N_T$), complex-to-complex and cell-to-cell variation implies that different parameters describe each short trajectory (length $N_t$) and therefore
the complexity is in the finite-sample-size limit.
In the current context, we simulate two experiments where the generative distributions are the (i) centered-Gaussian and the (ii) Laplace distributions, respectively. 

\subsubsection{Analysis}

The likelihoods are defined in Eqn.~\eqref{eqn:gausslaplace} for $\alpha = 1$ for the Laplace and $\alpha = 2$ for the centered-Gaussian model.
We define differences in the information criterion as the Gaussian minus the Laplace model, $\Delta (\cdot) \equiv (\cdot)_{2}-(\cdot)_{1}$, where negative values of $\Delta (\cdot)$ select the Gaussian model and positive values select the Laplace model.
The AIC complexity for both models is ${\cal K}_{\rm AIC}=1$ per trace. The QIC complexity is given by Eqn.~\eqref{eqn:alphacomp} per trace. Therefore the overall complexities for all traces are:
\begin{align}
{\cal K}_{\rm AIC} &= N_T, \\
{\cal K}_{\rm QIC} &= N_T\textstyle\frac{N_t}{N_t-\alpha}.
\end{align}
In this particular applications, QIC subsumes AICc, a previously proposed corrected AIC \citep{Hurvich1991,BurnhamBook}.
The average IC differences are:
\begin{center}
\resizebox{\columnwidth}{!}{
\scalebox{1.0}
{{\renewcommand{\arraystretch}{1.25}%
\begin{tabular}{lcc|cc|cc}
Generative &   \multicolumn{2}{l}{Sample size} & $\Delta\overline{ {\rm AIC}}$ & AIC Model & $\Delta\overline{{\rm QIC}}$ & QIC Model \\
 Model & $N_t$ & $N_T$ & (nats) &  Selection &   (nats) &  Selection\\
\hline
\hline
Gaussian & $5$ & 100 & $-46.7$ & Gaussian & $-5.1$  & Gaussian \\
\textbf{Laplace} & $5$ & 100 & $-17.1$  & \textbf{Gaussian} & $+24.5$ & \textbf{Laplacian} \\
\hline
Gaussian & $100$ & 5 & $-25.3$ & Gaussian & $-25.3$  & Gaussian \\
Laplace  & $100$ & 5 & $+32.6$  & Laplacian & $+32.6$ & Laplacian \\
\end{tabular}}}}
\end{center}
\medskip
\noindent
where we have highlighted the discrepancy in the analysis in bold. At large sample size ($N_t=100$), AIC and QIC both correctly select the generative distribution. But at small sample size, AIC incorrectly selects the Gaussian model when the generative model is Laplace. Qualitatively, the larger complexity of the Gaussian  relative to the Laplace model implies that the model has a  greater propensity to overfit by underestimating the information at the MLE parameters. As a result, AIC model selection prefers the Gaussian over the Laplace model, even when the Laplace model is both (i) the generative distribution and (ii) more predictive. Furthermore, since  $|\Delta {\rm AIC}|\gg 1$, the AIC analysis incorrectly indicates that there is extremely strong support for the Gaussian model.
In contrast to AIC, QIC selects the optimal model in both experiments and at both sample sizes.



%


\subsubsection{Comparison of QIC and cross-validation}

In the current example, the data is assumed to be \textit{unstructured} meaning that each observation $x_i$ is independent and identically distributed (in each trace). In these cases, there is a powerful alternative approach to estimating the predictive performance: Leave-One-Out-Cross-Validation (LOOCV). In the LOOCV estimate, each data point is predicted with parameters fit to the remaining $N-1$:
\begin{equation}
{\rm LOOCV}(x) = \sum_{i=1}^N h(x_i| \hat{\theta}_{x_{\ne i}}),
\end{equation}
where $x_{\ne i}$ is shorthand for the dataset excluding $x_i$.
To examine the relative performance LOOCV, AIC and QIC, we now consider performing model selection trajectory-by-trajectory ($N_T=1$) for five-step trajectories ($N_t=5$). For simplicity, consider data generated by the Laplace model where the complexity plays a central role in model selection due to the propensity of the Gaussian model to overfit. We then simulate the probability of the selection of the Laplace model by each criterion:

\begin{center}
\scalebox{1.0}
{{\renewcommand{\arraystretch}{1.25}%
\begin{tabular}{c|ccc}
Criterion & AIC & QIC & LOOCV \\
\hline
\hline
Probability of selecting Laplace & $34\%$ & $61\%$ & $53\%$ \\
\end{tabular}}}
\end{center}
\medskip
\noindent
which demonstrates that QIC outperforms both AIC and LOOCV, at least in the current context.

Why does LOOCV perform poorly? Although LOOCV is only weakly biased, it typically has a larger variance than QIC. To understand qualitatively why this is the case, we  define an effective LOOCV complexity:
\begin{align}
\Comp_\mathrm{CV}(x) \equiv \textstyle {\rm LOOCV}(x) -  h(x|\hat{\theta}_x),
\end{align}
which reinterprets LOOCV as a information criterion with a data-dependent complexity. 
The complexity $\Comp_\mathrm{CV}(x)$ acts like a weakly-biased estimator of the true complexity, but is subject to statistical variation, as shown in Fig.~\ref{LOOCV}. It is this variance that can lead to a loss in performance, even when the bias of the estimator is small. In contrast, the QIC complexity is constant in the current example. 

LOOCV and QIC each have respective advantages. The advantage of LOOCV is that  the data used to compute the estimated predictive performance were all generated by the true distribution. The frequentist complexity depends upon an assumed distribution, which can lead to a bias in QIC. 
LOOCV is also biased since it estimates the performance of predicting $1$ measurement given $N-1$ rather than $1$ measurement given $N$. Our own unpublished experiments indicate that whether LOOCV or QIC is more biased is  model and sample-size dependent. However, QIC does have two important and generic advantages: (i) it typically has less variance than LOOCV and (ii) it can also be applied to analyses of structured data where LOOCV cannot be applied, as illustrated in the next example.


\subsection{Anomalously large complexity and the Fourier regression model}
\label{Sec:Neutrino}
In this example we have two principal aims: (i) to explore the behavior of QIC in the context of a singular model with large complexity   and (ii) to demonstrate the dependence of the \QIC{} complexity on the model fitting algorithm. We present a model of simulated data inspired by the measurements of the seasonal dependence of the neutrino intensity detected at {\it Super-Kamiokande} \citep{Fukuda:2002uc}.

\subsubsection{Problem setup} We  simulate normally distributed intensities with arbitrary units (AU) with unit variance:
$\inliner{X_j \sim {\cal N}(\mu_j,1)}$,
where the true mean intensity $\mu_j$ depends on the discrete-time index $j$: 
\begin{eqnarray}
\mu_j  &=&  \sqrt{120+100\sin(2\pi j/N+\pi/6)}\ {\rm AU},
\end{eqnarray}
and the sample size is equal to the number of bins: $N=100$. This true distribution is therefore {\em unrealizable} for a finite number of Fourier modes. The generating model, simulated data and two model fits are shown in Figure \ref{Fig:Neutrino}, Panel A.

We expand the model mean ($\mu_i$) and observed intensity ($X_i$) in Fourier coefficients $\inliner{\tilde{\mu}_i}$ and $\inliner{\tilde{X}_i}$ respectively. A detailed description is provided in the Appendix Sec.~\ref{Sec:MoreNeutrino}. The MLE that minimizes the  information is $\inliner{\hat{\tilde{\mu}}_i = \tilde{X}_i.}$
We now introduce two different approaches to encoding our low-level model parameters $\{\tilde{\mu}_i\}_{i=-N/2...N/2}$: the {\it sequential} and {\it greedy algorithms}. In both cases, the models will be built by selecting a subset of the same underlying model parameters, the Fourier coefficients (${\tilde{\mu}}_i$).

\subsubsection{Sequential-algorithm analysis} In the {\it sequential  algorithm} we will represent our nested-parameter vector as follows: 
\begin{equation}
{    \theta}_{(n)} =  
\left( \begin{array}{cccc}
  &  \tilde{\mu}_{-1} & ... & \tilde{\mu}_{-n} \\
 \tilde{\mu}_0    &  \tilde{\mu}_1    & ... & \tilde{\mu}_n
\end{array} \right), \label{Eqn:sequentialModel}
\end{equation}
where all selected $\tilde{\mu}_i$ are set to their respective maximum likelihood values and all other $\tilde{\mu}_i$ are identically zero. We initialize the algorithm by encoding the data with parameters ${    \theta}_{(0)}$. We then execute a sequential nesting procedure, increasing temporal resolution by adding the Fourier coefficients $\tilde{\mu}_{\pm i}$ corresponding to the next smallest integer frequency index $i$. (Recall there are two Fourier coefficients at every frequency, labeled $\pm i$, except at $i=0$.) The cutoff frequency is indexed by $n$ and is determined by the model selection criterion.

\begin{figure*}
  \centering
    \includegraphics[width=1.0\textwidth]{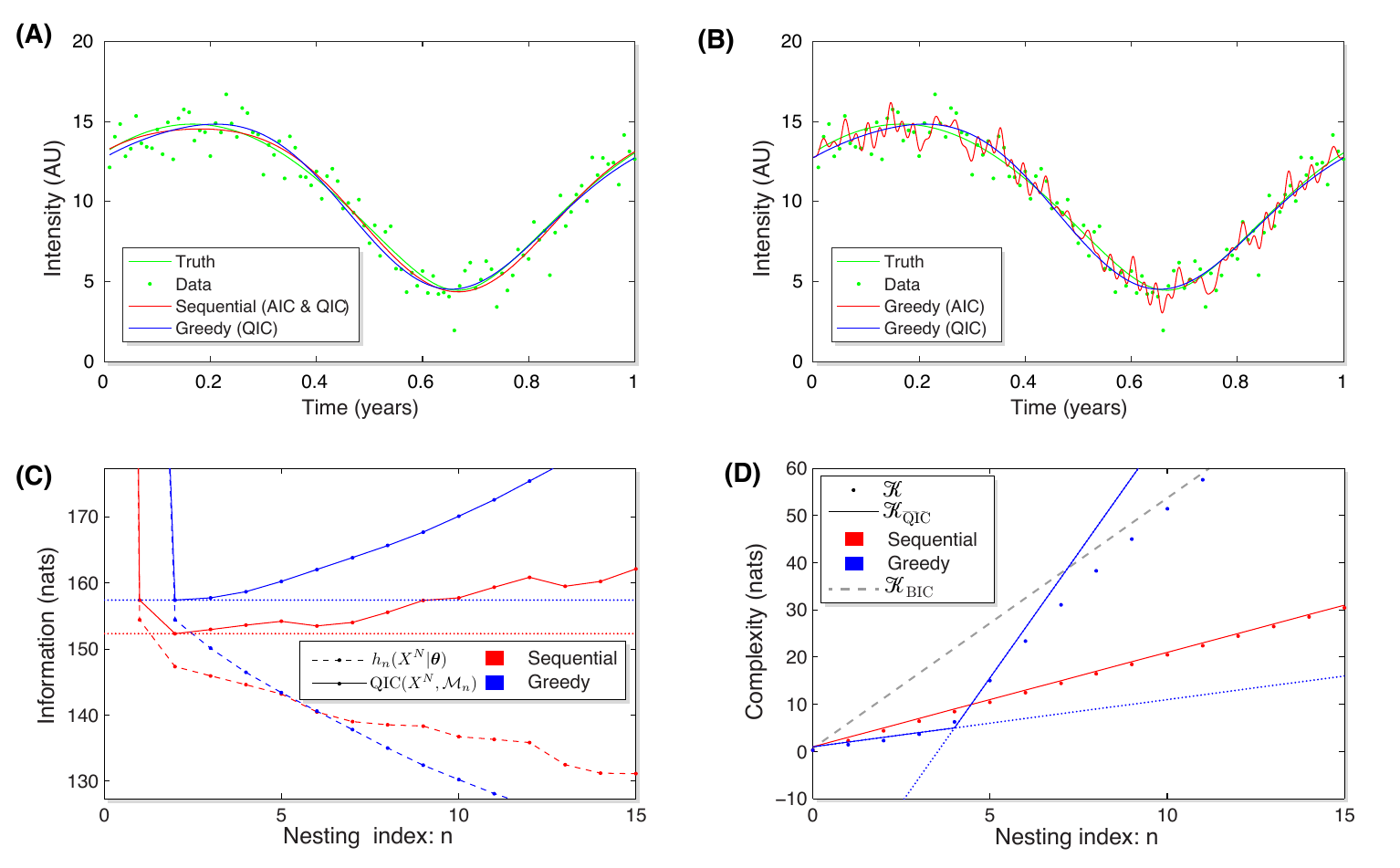}
      \caption{
     	 {\bf Panel A: Truth, data and models.} (Simulated for $N=100$.) The true mean intensity is plotted (solid green) as a function of season, along with the simulated observations (green points) and  models fitted using two different algorithms, sequential (red) and greedy (blue). 
 		{\bf Panel B: Failure of AIC for greedy algorithm.} (Simulated for $N=100$.) For the greedy algorithm, the coefficients selected using AIC (red) are contrasted with the coefficients chosen using QIC. The QIC mean estimates (blue) track the true means very closely. The AIC mean estimates (red) include many noise dominated Fourier modes. This model does not well represent the true seasonal behavior and would have poor predictive performance.
      {\bf Panel C: Information as a function of model dimension.}(Simulated for $N=100$.) The information is plotted as a function of the nesting index $n$. The true  information is compared with  the information for sequential (red) and greedy (blue) algorithm models. The dashed curves represent the information as a function of nesting index and both are monotonically decreasing. The solid curves (red and blue) represents the estimated average information (\QIC), which is equivalent to estimated model predictivity. The model selection criterion chooses the model size (nesting index) that is a minimum of \QIC. 
      {\bf Panel D: The true complexity matches \QIC{} estimates.} (Simulated for $N=1000$.) In the sequential-algorithm model, the true complexity (red dots) is  \AIC-like (solid red). In the greedy-algorithm model, the true complexity (blue dots) transitions from  \AIC-like (slope $= 1$) to \BIC{}-like (slope $\propto \log N$) at  $n=4$. In both cases, the true complexity is correctly predicted by \QIC{} (solid curve).  The \BIC{} complexity is a poor approximation of the predictive complexity in all models.
          \label{Fig:Neutrino}}
\end{figure*}

From the \AIC{} perspective, the complexity is simply a matter of counting the parameters fit for each model as a function of the nesting index. Counting the  parameters in Eqn.~\eqref{Eqn:sequentialModel} gives the expression for the complexity $\Comp_{\rm  \AIC} = 2n+1$,
since both an $\tilde{\mu}_i$ and an $\tilde{\mu}_{-i}$ are added at every level. Since this is a normal model with known variance, \QIC{}  estimates the same complexity as  \AIC. In the Bayesian analysis, the complexity is:
$\Comp_{\rm \BIC} = {\scriptstyle \frac{1}{2}}(2n+1)\log N$,
where $N=100$, which is significantly  larger than the  \AIC{} and \QIC{}. (See Sec.~\ref{Sec:BIC} for a discussion of the BIC analysis.)
Panel B of Figure \ref{Fig:Neutrino} shows \QIC{} model selection for the sequential algorithm. The $n=2$ nesting level minimizes QIC and this model ($n=2$) is shown in Panel A. The true and \QIC{} complexity are compared in Panel D for a sample size of $N=1000$. Both  \AIC{} and \QIC{} are excellent approximations of the true complexity.
%

%

\subsubsection{Greedy-algorithm analyis}  
Instead of starting with the lowest frequency and sequentially adding terms, an alternative approach would be to consider all the Fourier coefficients and select the largest magnitude coefficients to construct the model. In the {\it greedy  algorithm} we will represent the Fourier coefficients as 
\begin{equation}
 {    \theta}_{(n)} =  {\scriptstyle \left( \begin{array}{cccc}
0  &  i_1 & ... & i_n \\
 \tilde{\mu}_0    &  \tilde{\mu}_{i_1}    & ... & \tilde{\mu}_{i_n}
\end{array} \right)},
\label{eq:GreedyModel}
\end{equation}
where the first row represents the Fourier index and the second row is the corresponding Fourier coefficient. As before, all unspecified coefficients are set to zero.
We initialize the algorithm by encoding the data with parameters ${    \theta}_{(0)}$ and then we  execute a sequential nesting procedure: At each step in the nesting process, we choose the Fourier coefficient with the largest magnitude (not already included in ${    \theta}_{(n-1)}$).
The optimal nesting cutoff will be determined by model selection.

If one counts the parameters, the  \AIC{} and \BIC{} complexities are unchanged.
There are still two parameters in Eqn.~\eqref{eq:GreedyModel} at every nesting level $n$. For the \QIC{} complexity, the distinction between the sequential and greedy algorithms has profound consequences. The greedy-algorithm model is singular since the Fourier mode number $i_n$ becomes unidentifiable after the last resolvable Fourier mode is incorporated into $\inliner{{  \theta}_{(n)}}$. There are two approaches to computing the QIC complexity: (i)  Monte Carlo and (ii) an analytical piecewise approximation that we developed for computing an analogous complexity in change point analysis \citep{CPlong}. We will use the analytical approach, which gives a change in complexity on nesting of:
\begin{equation}
{\cal K}_n-{\cal K}_{n-1} \approx \begin{cases} 1, &-\Delta h>k\\ 
k, & {\rm otherwise} \end{cases}, \label{eqn:pw}
\end{equation}
where the change in information is defined $\inliner{\Delta h \equiv h_n(x|\hat{  \theta}_x)-h_{n-1}(x|\hat{  \theta}_x)}$ and the singular complexity is $k \approx 2 \log N$ (\textit{i.e.}~BIC scaling). The singular complexity $k$ arises due to picking the largest remaining Fourier mode. The approximation is given by computing the expectation of the largest of $N$ chi-squares, which is discussed in more detail in the supplement (Sec.~\ref{sec:fourierNesting}). If $-\Delta h>k$ the model is in a regular part of parameter spaces whereas if $-\Delta h<k$, the model is essentially singular \citep{CPlong}. The complexity is computed by re-summing Eqn.~\eqref{eqn:pw}.

\begin{figure*}
  \centering
   \includegraphics[width=1.0\textwidth]{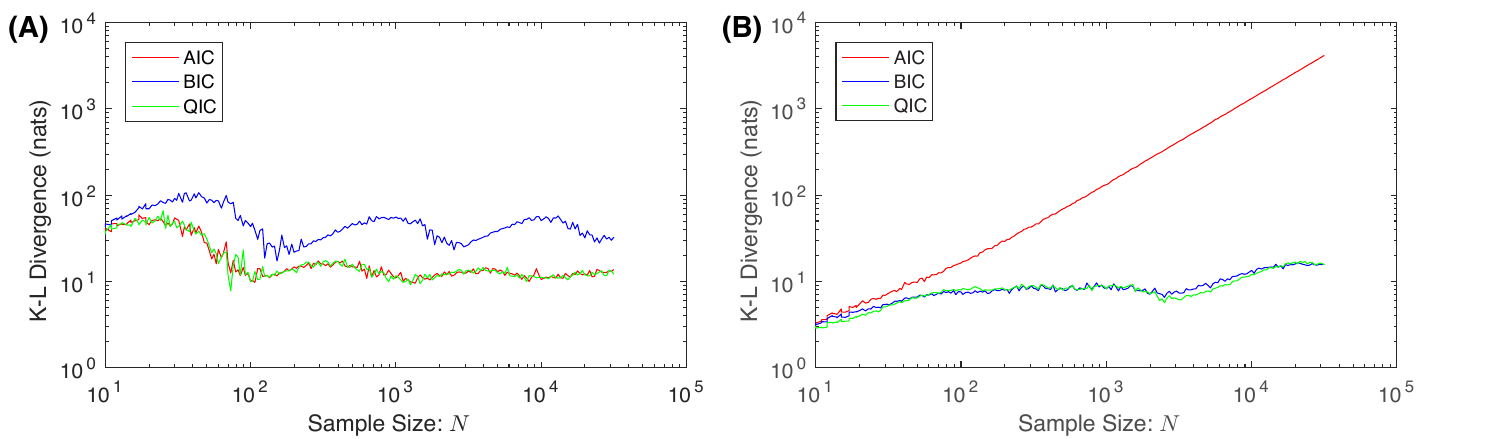} 
         \caption{
     {\bf Panel A: Performance of the sequential algorithm.} Simulated performance as measured by the KL Divergence $\overline{D}$ (Eqn.~\eqref{eqn:ModelKL}) of sequential algorithm at different sample sizes using AIC, QIC and BIC (lower is better). AIC and QIC are identical in this case; they differ only because of the finite number of Monte Carlo samples. Larger fluctuations are arise from the structure of true modes at the resolvable scale of a given sample size. 
	 {\bf Panel B: Performance of the greedy algorithm.} Simulated performance of greedy algorithm  as measured by the KL Divergence $\overline{D}$ (Eqn.~\eqref{eqn:ModelKL}) at different sample sizes using AIC, QIC and BIC (lower is better). QIC and BIC have very similar cutoff penalties. Because of the algorithmic sensitivity, QIC can have the appropriately complexity scaling with $N$ in both the greedy and sequential case.
          \label{Fig:performanceN}}
\end{figure*}

Panel B of Figure \ref{Fig:Neutrino} shows \QIC{} model selection for the greedy algorithm. The $n=2$ nesting 
level minimizes \QIC{} and this model ($n=2$) is shown in Panel A. The true and \QIC{} complexity  are compared in Panel D  for a sample size of $N=1000$. This large sample size emphasizes the difference between the slopes. In the greedy algorithm, only \QIC{} provides an accurate approximation of the true complexity. For large nesting index, the piecewise approximation made to compute the \QIC{} complexity fails due to order statistics. 
(The largest of $m$ $\inliner{\chi^2}$ random variables is larger than the second largest.) 
This is of little consequence since the complexity in this regime is not relevant to model selection.
The use of  \AIC{} model selection  in this context leads to significant over fitting by the erroneous inclusion of noise-dominated Fourier modes, as shown in Panel B of Fig.~\ref{Fig:Neutrino}. 

The predictive performance of the average selected model has been determined by Monte-Carlo simulations and is plotted in Fig.~\ref{Fig:performanceN}, for the greedy and sequential algorithms.
QIC shows correct scaling behavior for both fitting algorithms, which allows it to achieve good performance in both cases, whereas AIC (and not BIC) performs well in the Sequential case and BIC (and not AIC) performs well in the Greedy case.

%
%

\subsection{Anomalously small complexity and the exponential mixture model}
\label{Sec:expDecay}

In the greedy algorithm implementation of Fourier regression, both AIC and BIC {\em underestimated} the true complexity. But, the true complexity is not always underestimated by AIC. In sloppy models \citep{Machta:2013hl,Correspondence2018}, we find that the AIC (and BIC) approximation for the complexity  typically  {\em overestimate} the true complexity at finite sample size. To explore this phenomenon, we analyze an exponential mixture model.

In an exponential mixture model, $m$ different components decay at rate $\lambda_i$. The rates ($\lambda_i$), the relative weighting of each component in the mixture ($\omega_i$) and even the number of components ($m$) are all unknown. We represent the model parameters ${\bm \theta}=({\bm \lambda},\ {\bm \omega})$ and the candidate distribution function for the probability density of observing an event at time $t$ is: 
\begin{align}
q(t|{\bm \theta}) = \sum_{i=1}^m  \omega_i\ \lambda_i e^{-\lambda_i t},
\end{align}
with support $\omega_i,\lambda_i\in {\mathbb R}_+$ and constraint $\inliner{\sum_i \omega_i = 1}$. For $m>1$, this model is singular where $\omega_i=0$ or $\lambda_i=\lambda_j$ for $i\ne j$. Exponential mixture models are frequently applied in biological and medical contexts where the different rates might correspond to independent signaling pathways, or sub-populations in a collection of organisms, {\textit{etc}. 


\subsubsection{Problem setup} 
To explore the properties of the model, we simulate data from a realizable model with $m=4$ components and parameters:\begin{align}
{\bm \theta} \equiv \left(
\begin{array}{c}
{\bm \lambda} \\
{\bm \omega}
\end{array} \right) 
= \left(
\begin{array}{c c c c}
1 & 2 & 3 & 5 \\
0.3& 0.2 & 0.2 & 0.3
\end{array} \right).
\end{align}
For a large enough sample size, $N$, AIC could be expected to accurately estimate the complexity for an $m=4$ model. In practice, the sample size is always finite and therefore it is important to investigate the finite-sample size properties of the complexity. We simulated $N=100$ samples from the true distribution.

\subsubsection{Analysis}
In our statistical analysis, we consider just two competing models, $m=1$ and $4$ component models, for simplicity. For the AIC and BIC complexities, we used a model dimension of $K = 2m-1$ due to the normalization constraint on component weights $\omega_i$. The QIC complexity for $m=1$ has an analytic form given by Eqn.~\ref{eq:CompExp} while the complexity for $m=4$ was computed by Monte Carlo. 
The true complexity and the AIC, BIC and QIC approximations are compared for the two models below:
\begin{center}
\resizebox{\columnwidth}{!}{
\begin{tabular}{p{2cm}| p{2cm} p{2cm} p{2cm}  p{2cm} }
&  \multicolumn{4}{c}{Complexity ${\cal K}$ (nats)}   \\ 
Model	 & True	& QIC & AIC 	& BIC 	 	 \\
\hline
\hline
$m = 1$  & 1.77		& 1.01 & 1 	& 4.61 	 	 \\ 
$m = 4$  & 3.33   	& 3.45 & 7 	& 16.1      \\
\end{tabular}}
\end{center}
QIC shows excellent agreement with the true complexity for $m=4$.  The discrepancy when $m= 1$ occurs because QIC has approximated the true distribution ($m=4$) with the fitted model ($m=1$). The true distribution in this case is {\it not realizable}, but nonetheless this approximation still provides the best estimate of the true complexity.
For the one component model ($m=1$), AIC makes nearly the same estimate for the complexity as QIC, but it significantly overestimates the complexity of the larger 4 component model $(m=4)$. At finite sample size, this model is therefore more predictive than estimated by AIC. The BIC complexity never accurately approximates the true complexity.

The difference in estimated complexity has important consequences for model selection. We will define the difference  $\inliner{\Delta (\cdot) \equiv (\cdot)_1-(\cdot)_4}$, where $\inliner{\Delta (\cdot)>0}$ implies the $m=4$ model is expected to be more predictive.  
Consider the training-sample average differences between the MLE information, the information criteria and the cross entropy difference:
\begin{center}
\resizebox{\columnwidth}{!}{
\resizebox{\columnwidth*12/14}{!}{
\begin{tabular}{ p{2cm} p{2cm} p{2cm} p{2cm} p{2cm} }
  \multicolumn{5}{c}{Average information difference (nats)}  \\ 
   $\Delta \overline{H}$ & $\Delta \overline{{\rm QIC}}$ &  $\Delta \overline{h\inliner{(X|\inliner{\hat{\bm \theta}}_X)}}$ & $\Delta \overline{{\rm AIC}}$ 	& $\Delta \overline{{\rm BIC}}$ 	  \\
\hline
\hline
 3.73  & 2.84  &   5.29 	& -0.72 	& -8.53 	
\end{tabular}}}
\end{center}
In a nested model, the larger model is always favored by $\Delta h$ due to overfitting. The average cross entropy is also positive, which implies that the trained $m=4$ model is more predictive than the $m=1$ trained model on average.
The QIC complexity most-closely estimates the true complexity and there is the best agreement between the average cross entropy difference and average QIC. QIC also favors the $m=4$ model.
Due to the overestimate of the complexity for $m=4$,  both AIC and BIC tend to favor the smaller model. 

Although QIC better estimates the true complexity on average, unlike the AIC and BIC estimates, it depends on the MLE parameter estimate and so there are statistical fluctuations in the estimated complexity. A large variance  might still lead to a degradation in model selection performance, even if the mean  were unbiased. We therefore compute the model selection probabilities and the expected predictive performance of model the selection criteria for AIC, BIC and QIC by computing KL Divergence, averaged over the training set:
\begin{center}
\resizebox{\columnwidth}{!}{
\begin{tabular}{p{4cm} |  p{2cm} p{2cm} p{2cm} }
Performance     &  \multicolumn{2}{c}{Model selection criterion }  \\ 
metric			&  AIC 	& BIC 	& QIC  	 \\
\hline
\hline
${\rm Pr}_1$ &  0.64 & 0.98 & 0.19   \\
${\rm Pr}_4$ &  0.36 & 0.02 & 0.81 \\
\hline
$\overline{D}$ (nats)\ \ (Eqn.~\eqref{eqn:ModelKL}) 	&  4.02 	& 5.24 	& 2.17\\
\end{tabular}}
\end{center}
where ${\rm Pr}_{\cal m}$ is the probability of selecting model ${\cal m}$, \textit{Choose ${\cal m}$} is a criterion where model ${\cal m}$ is always chosen.
As expected, QIC has superior performance to AIC and BIC since it picks the $m=4$ model with higher probability.
Both AIC and BIC underestimate the performance of the larger model and therefore need a significantly larger dataset to justify the selection of the model family that contains the true distribution. We believe this example is representative of many systems biology problems where the complexity is significantly smaller than predicted by the model dimension alone.

\section{Discussion}  

Although the \AIC{} and \BIC{} complexities depend only on the number of parameters,  the true and QIC complexities depend on the likelihood and the fitting algorithm itself.  In general, the QIC complexity will not be exactly equal to the true complexity and therefore QIC remains a biased estimator of cross entropy. 
 In this section, we shall outline the known properties of QIC.

%
%
%

\subsection{$\QIC{}$ subsumes extends both AIC and  AICc}

\label{Sec:ext}

In comparing QIC to existing information criteria, it is first important to note that, for an important class of analyses, QIC is expected to be exactly equivalent to AIC or corrected AIC. In the large sample size limit of regular models, the frequentist complexity is equal to the AIC complexity and therefore AIC and QIC are identical. Furthermore, 
$\QIC{}$ subsumes an important class of previously proposed refinements to AIC. 
These complexities follow from the assumption of realizability, and the special case of parameter-invariant frequentist complexity discussed in \ref{sec:constantComp}.
The AIC complexity is itself exact for the normal model with unknown mean(s) and known variance at any sample size.
Another example is $\AIC_\textrm{C}$, derived in the context of linear least-squares regression with unknown  variance. In this case the complexity is \citep{Hurvich1989}:
\begin{equation}
\Comp = \dimK \textstyle \frac{N  }{N - \dimK - 1} \label{eq:CompVar},
\end{equation}
which is equal to $K$ in the large-sample-size limit ($N\rightarrow \infty$), but deviates significantly for small $N$ corrected AIC \citep{Hurvich1991,BurnhamBook}. 
Another exact result is found for the exponential model, $\inliner{q(x|\theta) = \theta\, e^{-\theta x}}$, where the complexity is \citep{BurnhamBook}
\begin{equation}
\Comp = \textstyle \frac{ N}{N - 1} \label{eq:CompExp}.
\end{equation}
The appealing property of these complexities is that, like AIC, they do not require knowledge of the true distribution and therefore maintain all the advantages of AIC while potentially correcting for finite-sample-size effects.

Burnham and Anderson have previously advocated the use of Eqn.~\eqref{eq:CompVar} even outside the case for linear regression, on the grounds that some finite-sample-size correction is better than none \citep{BurnhamBook}. The QIC complexity is a more principled approach, using the assumption of realizability without presupposing a complexity of the model.  When the frequentist complexity of a particular model {\em is} constant, \QIC{}  recovers a form of $\AIC_\textrm{C}$. When it is not, the generative parameters must be estimated using the frequentist complexity, Eqn.~\eqref{eqn:QICcomp}.

\subsection{Asymptotic bias of the QIC complexity}

A canonical approach to analyzing the performance of an estimator is to study the bias of that estimator in the large-sample-size limit. An asymptotic unbiased estimator of the cross-entropy will be an asymptotically efficient model selection criteria under standard conditions (See \citet{Arlot2010,Shao1997} for details).  Efficiency is an important goal for predictive model selection\citep{Shao1997,Yang2005,birge2007minimal}.

QIC is not a significant improvement over AIC in terms of asymptotic bias.  First, just as with AIC, we must assume that the true model is realizable (although this condition can be relaxed, see \ref{sec:realizability}) .
If the true model is realizable $\phi \in \Theta$, then we can Taylor expand the frequentist complexity around the true parameter value:
\begin{align}
\label{eqn:TaylorExpansion}
\overline{\Comp(\phi+\delta \theta_X)}  =  \Comp(\phi) + \overline{\delta \theta}_X \cdot \nabla \Comp(\phi) +   \textstyle\frac{1}{2}\overline{\delta \theta_X \otimes \delta \theta_X}\cdot \nabla\otimes\nabla\Comp(\phi) + \hdots, 
\end{align}
where the over line represents expectations with respect to $X\sim q(\cdot|\phi)$. If the estimated parameters are unbiased, the second term is zero.  For nonsingular points the third term is asymptotically zero---but at non-singular points QIC is asymptotically equal to AIC.  At singular points the bias due to the third term is expected to be greater than ${\cal O}(N^{-1})$ and QIC {\em will} be asymptotically biased. 

However, in practice the QIC estimate of the complexity often appears to be \textit{good enough}, and certainly superior to the alternatives. For example, the Greedy algorithm of the Fourier analysis is a useful test case. This problem is singular. The use of the AIC complexity in this problem leads to a catastrophic breakdown in model selection: The number of overfit parameters added is very large and grows with the sample size $N$. In contrast, the QIC estimate of the complexity, though biased, has the correct $\log N$ scaling behavior near the singular point: the QIC method shows excellent model selection performance in this context. 

We measured the relative performance of QIC using three metrics: we compared (i) QIC complexity to the true complexity and (ii) QIC to the cross entropy, and (iii) directly computing the KL Divergence of the trained-selected model. By all three metrics, we demonstrate that QIC outperforms AIC and BIC. We therefore conclude that, while QIC does not generically offer asymptotic efficiency when AIC does not, QIC is often vastly superior to AIC at a finite sample size, where all real analyses occur.

\subsection{Advantages of QIC}
QIC has several advantages compared with existing methods.  Although QIC is not universally unbiased, a good estimator should balance bias and variance---in a bias-variance tradeoff \citep{geman1992neural,piironen2017comparison}.  QIC tends to have both relatively low bias (compared to AIC, $\mathrm{C}_p$ and similar penalized methods) and low variance, compared to CV, bootstrap, and the Takeuchi information criterion (TIC) \cite{bozdogan2000akaike}.

\subsubsection{QIC has smaller biases than AIC and similar methods}
Although QIC and AIC have similar asymptotic behavior and performance, at finite sample size, AIC will have greater bias in a cross entropy estimator, and will typically have greater predictive loss.  This performance loss due to the bias of AIC can be significant \cite{barron1999risk,birge2007minimal,efron2004estimation}, especially for small $N/\Comp$.  For regular, realizable models with constant or slowly varying $\Comp(\theta)$, QIC will have negligible bias even at small sample size.

\subsubsection{QIC has smaller variance than empirical methods}

One practical method to circumvent the QIC assumption of realizability is the use of  estimators  depend only on empirical expectations taken with respect to the observed data (i.e.~LOOCV, bootstrap, etc).
%
%
Empirical estimates for the complexity such as the bootstrap methods are guaranteed to be asymptotically unbiased in a very wide range of model selection scenarios.  If the sample size is large, cross-validation has highly desirable properties. However empirical methods are inferior to both AIC and AICc in the regular limit because they suffer from a large variance resulting from the subsampling procedure \citep{Shibata1997,BurnhamBook,shibata1981optimal,efron2004estimation,birge2007minimal}.  This increased variance leads to degraded performance when unbiased estimators of the complexity are available.  QIC therefore has provably superior performance in many situations \citep{efron2004estimation}.

\subsubsection{QIC is applicable to models of structured data such as time series}
Both LOOCV and bootstrap rely on an assumption that the data are unstructured, i.e. they take the form of independent and identically distributed random vectors.
QIC can be applied, without modification, to structured data such as time series, where correlations exist between measurements.  We originally developed a version of QIC in one such structured context: the change-point problem \citep{CPlong}.  If calculations of QIC requires a Monte-Carlo calculation, data are sampled from the joint distribution, which therefore preserves the relevant dependencies in the data. In contrast, it is not as straightforward to \textit{leave out} or \textit{subsample} a data-point when doing Fourier analysis or DNA sequencing, although workarounds exist in specialized situations (e.g. generalized CV \citep{craven1978smoothing}).

\subsubsection{QIC responds to the effects of manifold geometry}
QIC is non-perturbative, unlike AIC and TIC, and other methods that rely on Taylor expansion. The putative distribution of $\hat{\theta}_x$ in the frequentist expectation will explore parameter space in the vicinity of the optimal value, and meet constraints and nearby singularities.  Although these features usually result in QIC being biased, these biases are often small when compared with the complete failure of other methods. Two of our example applications are in singular spaces, where empirical evidence suggests that QIC is  robust with complexity estimates that are accurate enough to achieve good performance.

\subsubsection{QIC can account for the multiplicity.} Assuming a generative model gives QIC the ability to simulate the behavior of the entire procedure including stopping rules, outlier removal, thresholding and the fitting algorithm itself. In particular, the order in which a model family is traversed can have a profound effect on the complexity due to the multiplicity of competing models \citep{genovese2006false,draglia1999multihypothesis}. These multiplicity effects are ubiquitous, and in frequentist tests they lead to Bonferonni corrections \citep{bland1995multiple,hochberg1988sharper} to the significance level. QIC automatically generates an information-based realization of the Bonferonni corrections---models with large multiplicity have substantially increased complexities.  This increase in complexity lead to a much stronger preference for smaller models in the presence of multiplicity than in sequential model selection. We studied these effects in Sec.~\ref{Sec:Neutrino}.

\subsubsection{QIC accounts for the learning algorithm}

Algorithmic dependence plays an interesting and important role in determining the complexity in some simple applications we discuss. The two approaches to the neutrino problem illustrate this point: Although both the sequential and greedy algorithms represent the intensity signal as Fourier modes, the complexities are fundamentally different as a result of the fitting algorithms. This algorithmic dependence is typical. For example, the greedy addition of regressors in linear regression problems is a common realization of a singular model that results in significant increases in complexity. QIC facilitates an information-based approach to  these problems for the first time and reinforces the notion that the fitting algorithm can be of equal importance to the number of model parameters.

\subsection{Conclusion} 
We have proposed a new information criterion: the Frequentist Information Criterion (QIC). 
QIC is a significantly better approximation for the true complexity and results in better model selection performance than AIC in many typical analyses.
Although, QIC is equal to AIC in the large-sample-size limit of regular models, 
QIC is a superior approximation in regular models at finite sample size as well as singular model at all sample sizes and can account changes in the complexity due to algorithmic dependence.
The QIC approach to model selection is objective and free from {\it ad hoc} prior probability distributions, regularizations, and the choice of a null hypothesis or confidence level. 
It therefore offers a promising alternative to other model selection approaches, especially when existing information-based approaches fail.

\appendix

\section{Appendix}
\subsection{Exponential families}
\label{app:bregmann}
An important case is the exponential-family, where the likelihood can be written:
\begin{align}
q(x|\theta) = \exp[t(x)\cdot \theta - N \psi (\theta) + r(x)],
\end{align}
the sufficient statistics $t(x)$ and function $r(x)$ are functions of the dataset $x$ only and $\psi (\theta)$ is a function of the parameters only and $N$ is the sample size. In this case, the complexity can be computed from Eqn.~\eqref{eq:comp} and can be written:
\begin{align}
{\cal K}(\theta ) =  \expect{X,Y|\theta} \left[t(X)-t(Y)\right]\cdot \hat{\theta}_X, \label{eqn:compexp}
\end{align}
where $X$ and $Y$ are two independent datasets of sample size $N$ generated from distribution $q(\cdot|\theta)$ and
\begin{align}
\hat{\theta}_X \equiv (\nabla \psi)^{-1}[t(X)/N],
\end{align}
where $(\nabla \psi)^{-1}$ is the functional inverse of the gradient of function $\psi$.

\subsection{Modified-Centered-Gaussian distribution}

\label{sec:modgauss}
The likelihood for the Modified-Centered-Gaussian model is given by Eqn.~\eqref{eqn:gausslaplace}. The MLE parameters and sufficient statistic are:
\begin{align}
\hat{\theta}_x &= - \frac{N}{\alpha\, t(X)}\\ 
t(X) &= -\sum_{i=1}^N |x|^\alpha,
\end{align}
respectively. The sufficient statistic $t$ is distributed like a Gamma distribution:
\begin{equation}
-t \sim \Gamma( N/\alpha, \lambda),
\end{equation} 
which has well-known moments:
\begin{equation}
\overline{(-t)^m} = \textstyle \frac{\Gamma(m+N/\alpha)}{(-\lambda)^m\Gamma(N/\alpha)}.
\end{equation}
Using the last results in combination with expression for the complexity of an exponential model, Eqn.~\eqref{eqn:compexp}, we find:
\begin{equation}
{\cal K} = \textstyle\frac{N}{N-\alpha}\ \ \ \ {\rm for}\ \ \ \ \ N>\alpha,
\end{equation}
which is always larger than the AIC complexity $K = 1$ for $\alpha>0$.

\subsection{The component selection model } 
\label{sec:compSel}
For convenience, consider a true model where $j = n$, which is general due to permutation symmetry. Let the observations be defined as:
\begin{equation}
X_j = \xi_j+[j=n]\mu,
\end{equation} 
where we have used the Iverson bracket and the  $\xi_j$ are iid random variables centered around zero with unit variance. The MLE parameters for the model are:
\begin{align}
\hat{\imath} &= \arg \max_j X_j^2,\\
\hat{\mu} &= X_{\hat{\imath}}.
\end{align}
The complexity can then be written:
\begin{equation}
{\cal K}(\theta) = \expect{\xi}\left\{\max_jX^2_j-\mu^2[\hat{\imath}=n]\right\},
\end{equation}
which can be computed using one-dimensional integrals of the CDFs.

It is useful to consider the large and small multiplicity limit. For large multiplicity ($n$), the complexity is
\begin{equation}
{\cal K}(\theta) = 2\log n - \log \log n-2 \log \Gamma(\textstyle\frac{1}{2}) + 2 \gamma + ...,\label{eqn:maxtheta}
\end{equation}
where $\gamma$ is the EulerÐMascheroni constant \citep{Hashorva2012}. For $n = 1$ or sufficiently large $\mu$, there is no multiplicity and we recover the AIC result:
\begin{equation}
{\cal K}(\theta) = 1.
\end{equation}

\subsection{n-cone}
\label{sec:conecalc}
Following notation used in special relativity, we denote the \textit{space-like} component of a vector $\vec{A} = \left\{A_2,\hdots A_{n} \right\}$ and the \textit{time-like} component $A_1$. The implicit function of constraint is
\begin{align}
\rho(\theta) = \vec{\mu}^2 - (c \mu_1)^2 = 0,
\end{align}
which is to say that the mean must lie on the \textit{light cone}. The observations $X = (X_1 , \vec{X})$ can be represented as:
\begin{equation}
X = \mu + \xi,
\end{equation}
where $\xi$ is an $n$-vector of iid random variables normally around zero with unit variance.
The MLE parameters satisfying the constraints are:
\begin{align}
\vec{\hat{\mu}} &=\frac{ \left( \frac{c \left|X_1 \right| }{\left|\vec{X} \right|} +c^2 \right)}{c^2+1}\vec{X} ,\\
\hat{\mu}_1 &= \frac{X_1 + c\, \text{sgn}(X_1) \left|\vec{X} \right|}{c^2+1}.
\end{align}
We can take the expectation using known properties of the non-central $\chi$ distribution.  The result can be expressed in terms of the generalized Laguerre polynomials:
\begin{align}
{\cal K}(\theta) &= \frac{c^2 (k-1)-c \vec{\mu}^2 \left(\sqrt{\frac{\pi }{2}} \mu_1 \, \text{erf}\left(\frac{\mu_1}{\sqrt{2}}\right)+e^{-\frac{\mu_1^2}{2}}\right) L_{-\frac{1}{2}}^{\frac{k-1}{2}}\left(-\frac{\vec{\mu}^2}{2}\right)}{c^2+1} \\
& \qquad+ \frac{c \left(\sqrt{\frac{\pi }{2}} \mu_1 \, \text{erf}\left(\frac{\mu_1}{\sqrt{2}}\right)+2 e^{-\frac{\mu_1^2}{2}}\right) L_{\frac{1}{2}}^{\frac{k-3}{2}}\left(-\frac{\vec{\mu}^2}{2}\right)+1}{c^2+1}.
\end{align}
This result recovers the known results of AIC on the realizable surface far from the singularity, and ${\cal K} = 1$ when c is very large, corresponding to a needle-like geometry where the surface of constraint is essentially one-dimensional compared to the scale of the Fisher information. 
%
%

\subsection{Fourier Regression nesting complexity}
 \label{sec:fourierNesting}
A literal treatment of the QIC algorithm requires a Monte Carlo simulation. However, as can be seen in Fig.~1, this complexity interpolates between two limiting  behaviors that can be treated analytically. To treat the nesting complexity analytically, we will make two assumptions: (i) All previously included models are unambiguously resolved and (ii) the number of modes included is small compared to the total $n$. Under these two assumptions, the nesting complexity is equivalent to selecting the largest magnitude coefficient of the remaining unselected Fourier components. Since each is independent and normally distributed, this problem is exactly equivalent to a problem that we have already analyzed: the component selection model.  
In this case, we can simply reuse the complexity derived in Eqn.~\eqref{eqn:maxtheta} as the nesting complexity, with limiting behavior:
\begin{align}
{k_i}(\theta) = 
\begin{cases}
2 \log N \quad &\text{when $\mu^2  \ll 2\log N$}\\
1 \quad  &\text{when $\mu^2  \gg 2\log N$}\\
\end{cases},
\end{align}
in exact analogy to Eqn.~\eqref{eqn:maxtheta} where the number of components $n = N$. The total complexity can be summed,
\begin{align}
\Comp(\theta) = \sum_i k_i (\theta).
\end{align}
We have previously used this approximation in the context of change-point analysis \citep{CPlong,CPshort}.

\subsection{$L_1$ Constraint}
\label{app:L_1}
We use the simplex projection algorithm described in  \citet{duchi2008efficient} with the MATLAB code to project onto an $L_1$ ball provided by John Duchi at \url{https://stanford.edu/~jduchi/projects/DuchiShSiCh08.html}.  We computed the complexity using $10^5$ samples on a $10^{-1}$ grid, with the resulting complexity linearly filtered in Fourier space.

\subsection{Curvature and QIC unbiasedness under non-realizability}
\label{sec:realizability}
If $|\theta_X - \theta_0|$ is small (on average) relative to the inverse-mean-curvatures of the manifold ${\bm \Theta}$, then we have that the true complexity is given by
\begin{align}
\label{eq:realizable}
\Comp(\phi) &\approx  \expect{X|\phi}  \left\{\KLD(\theta_0 || \theta_X) - \KLstat_X(\theta_0 || \theta_X) \right\} \\
\intertext{This follows from Amari's ``generalized pythagorean theorem" \citep{amari2007methods,Amari1985} where $\KLD(\phi|| \theta)$ is analogous to the half-squared-distance between $\phi$ and $\theta$. If $\hat{\theta}_X$ is a MLE then $\KLstat_X(\theta_0 || \theta_X)$ is equivalent to another K-L divergence \cite{amari2007methods}. We can finally write this as}
&\approx \expect{\theta_X|\phi}  \left\{\KLD(\theta_0 || \theta_X) + \KLD(\theta_X || \theta_0) \right\}.
\end{align}

For (nearly) flat manifolds, such as the unconstrained exponential family, with $\hat{\theta}_X$ being the MLE, we do not need the distribution of the {\em data} $X|\phi$ to be well approximated by $X|\theta_0$, we only need the distribution of the {\em fitted parameters} to match $\expect{\theta_X|\phi} \approx  \expect{\theta_X|\theta_0}$.
\begin{align}
\Comp(\phi) &\approx \expect{\theta_X|\theta_0}  \left\{\KLD(\theta_0 || \theta_X) + \KLD(\theta_X || \theta_0) \right\}\\
 &= \Comp(\theta_0)
\end{align}
In which case the model is {\em effectively} realizable for our purposes in the sense that $\Comp(\theta_0)$ is unbiased, even though $ \KLD(\phi || \theta_0)$ may be large. Eqn.~\eqref{eqn:TaylorExpansion} and the subsequent considerations then apply.

We would expect QIC to be biased if $\theta_0$ poorly describes the variance of $\theta_X$.  For instance, if we assume a fixed, incorrect, variance $\sigma'^2$, instead of the true value of $\sigma$, this will bias the QIC complexity by a scale factor of $\sigma^2/\sigma'^2$.  Although  we'd expect Eq.~\ref{eq:realizable} to be very generally asymptotically true, our complexity landscapes show that the presence or absence of extrinsic curvature of $\Theta$ is an important factor in whether or not the variance of $\theta_X$ will be well estimated by $\theta_0$. When the true distribution is not realizable, the variance of $\theta_X$ will depend on the curvature, and QIC may have significant bias.

\idea{Approximations for marginal likelihood} 

\label{Sec:BIC}

A second canonical information criterion (BIC) is motivated by Bayesian statistics. In Bayesian model selection, the canonical approach is to select  the model with the largest marginal likelihood:
\begin{equation}
q(x) \equiv \int_{\Theta}\!\! {\rm d}{\bm \theta}\ \varpi({\bm \theta})\ q(x|{\bm \theta}),
\end{equation} 
where $\varpi$ is the prior probability density of parameters ${\bm \theta}$. If we assume (i) the large $N$ limit, (ii) that the model is regular, (iii) the model  dimension is constant as $N$  increases and (iv) the prior is uninformative, the negative log of the marginal likelihood can be computed using the Laplace approximation \citep{Schwarz1978a,BurnhamBook}:
\begin{equation} 
-\log q(x) = h(x|\hat{\bm \theta}_x) + {\textstyle \frac{1}{2}} K\log N + \log \textstyle \frac{ \sqrt{ (2\pi)^K \det {\bm I}}}{\varpi(\hat{\bm \theta}_x)}+... 
\end{equation}
where $K$ is the dimension of the model and  ${\bm I}$ is Fisher Information Matrix. The first three terms have $N^1$, $\log N$ and $N^0$ scaling with sample size $N$, respectively. A canonical approach is to keep only the first two terms of the negative log of the marginal likelihood, which define the Bayesian Information Criterion (BIC):
\begin{equation}
{\rm BIC}(x) = h(x|\MLE_x)+{\textstyle \frac{1}{2}} K\log N, \label{eqn:BIC}
\end{equation}
which has the convenient property of dropping the prior dependence since it is constant order in $N$ \citep{Schwarz1978a,BurnhamBook}. 
The BIC complexity grows with sample size and is therefore larger than the AIC complexity  in the large $N$ limit. This tends to lead to the selection of smaller models than AIC. 
Since the prior typically depends on \textit{ad hoc} assumptions about the system, the absence of   prior dependence is an attractive feature of BIC. On-the-other-hand, in many practical analyses $\log N$ is not large, which makes the canonical interpretation of BIC dubious. A more palatable interpretation of BIC is to imagine withholding a minimal subset of the data (\textit{i.e.}~$N \approx 1$) to generate an informative prior, then computing marginal likelihood. This sensible Bayesian  procedure is well approximated by BIC \citep{lamont2016lindley}.   



\subsection{Seasonal dependence of the neutrino intensity }

\label{Sec:MoreNeutrino}

\subsubsection{Analysis of the data} We expand the model mean ($\mu_i$) and observed intensity ($X_i$) in Fourier coefficients $\tilde{\mu}_i$ and $\tilde{X}_i$ respectively. The MLE parameters that minimize the  information are $\hat{\tilde{\mu}}_i = \tilde{X}_i.$
We now introduce two different approaches to encoding our low-level model parameters $\{\tilde{\mu}_i\}_{i=-N/2...N/2}$: The {\it Sequential} and {\it Greedy Algorithms}. Note that in both cases, the models will be

\subsubsection{Analysis of the data} We expand the model mean ($\mu_i$) and observed intensity ($x_i$) into Fourier coefficients $\tilde{\mu}_i$ and $\tilde{X}_i$ respectively:
\begin{eqnarray}
\mu_j &=& \sum_{i=-N/2}^{N/2} \tilde{\mu}_i \psi_i(j)\ \  \ \ {\rm where }\ \ \ \   \tilde{\mu}_i = \sum_{j=1}^N \mu_j \psi_i(j),  \\
x_j &=& \sum_{i=-N/2}^{N/2} \tilde{X}_i \psi_i(j)\ \  \ \ {\rm where }\ \ \ \   \tilde{X}_i = \sum_{j=1}^N x_j \psi_i(j), 
\end{eqnarray}
where the orthonormal Fourier basis functions are defined: 
\begin{equation}
\psi_i(j) \equiv  N^{-1/2}\begin{cases}
\sqrt{2}\, \cos( 2\pi ij/N ), & i<0\\
1, & i=0\\
\sqrt{2}\, \sin( 2\pi ij/N ), & i>0.
 \end{cases}
\end{equation}
Substituting these expressions into the expression of the data-encoding information gives
\begin{equation}
h(X^N|{\bm \theta}) = { \frac{N}{2}}\log 2\pi \sigma^2 + \frac{1}{2\sigma^2} \sum_{i=-N/2}^{N/2} (\tilde{X}_i-\tilde{\mu}_i)^2, \label{Eqn:infofourierexpansion}
\end{equation}
where we have used the orthagonality in the large $N$ limit for all terms. We chose the eigenfunction normalization in order to give this expression its concise form.

Note that there is no need to (re)compute the information \textit{etc} since the structure of the problem is identical to the resonance problem discussed above.

\begin{acknowledgements}
P.A.W.~and C.H.L.~would like to thank  J.~Kinney, K.~Burnham, S.~Presse, and M.~Drton for advice and discussions and M.~Lind\'en, N.~Kuwada, and J.~Cass for advice on the manuscript.
\end{acknowledgements}

\bibliographystyle{apalike}      
\bibliography{ficbib}   

\end{document}